\documentclass[a4paper, 10pt, twocolumn, twoside]{article} 

\usepackage{00_Template/ISARC}

\usepackage{lscape}
\usepackage{hologo}
\usepackage{xcolor} 

\usepackage{siunitx} 
\DeclareSIUnit{\nothing}{\relax} 

\usepackage{import} 

\usepackage{multirow}	
\usepackage{booktabs}

\DeclareAcronym{GeoLab}{
    short = GeoLab,
    long  = Geodätisches Labor,
    long-genitive-form = Geodätischen Labors,
    long-dative-form = Geodätischen Labor,
    }
    
\DeclareAcronym{2D}{
	short   = 2D,
	long    = Zweidimensional,
	}
	
\DeclareAcronym{3D}{
	short   = 3D,
	long    = Dreidimensional,
	}

\DeclareAcronym{BIM}{
	short   = BIM,
	long    = Building Information Modeling,
	}
	
\DeclareAcronym{BIM model}{
	short   = BIM model,
	long    = building information model,
	}
	
\DeclareAcronym{URDF}{
	short   = URDF,
	long    = Universal Robot Description Format,
	}
	
\DeclareAcronym{SOTA}{
	short   = SOTA,
	long    = State Of The Art,
	}	

\DeclareAcronym{SDF}{
	short   = SDF,
	long    = Simulation Definition Format,
	}
	
\DeclareAcronym{SVG}{
	short   = SVG,
	long    = Scalable Vector Graphics,
	}
	
\DeclareAcronym{BIRS}{
	short   = BIRS,
	long    = Building Information Robotic System,
	}

\DeclareAcronym{Scan-vs-BIM}{
	short   = Scan-vs-BIM,
	long    = comparison between a point cloud and a \ac{BIM} model,
	}

\DeclareAcronym{GIS}{
	short   = GIS,
	long    = Geographic Information System,
	}

\DeclareAcronym{BGU}{
	short   = BGU,
	long    = \DAfaculty{},
	}
    
\DeclareAcronym{EMM}{
	short   = EMM,
	long    = Environment Mapping Module,
	}
	
\DeclareAcronym{SLAM}{
	short   = SLAM,
	long    = Simultaneous Localization and Mapping,
	}
	
\DeclareAcronym{V-SLAM}{
	short   = V-SLAM,
	long    = Visual SLAM,
	}
	
\DeclareAcronym{UGV}{
	short   = UGV,
	long    = Unmanned Ground Vehicle ,
	}	
	
\DeclareAcronym{UAV}{
	short   = UAV,
	long    = Unmanned Aerial Vehicle,
	}	
		
 \DeclareAcronym{UV}{
	short   = UV,
	long    = Unmanned Vehicle,
	}	
	
\DeclareAcronym{UVs}{
	short   = UVs,
	long    = Unmanned Vehicles,
	}
	
\DeclareAcronym{LiDAR}{
	short   = LiDAR,
	long    = Light Detection and Ranging,
	}

\DeclareAcronym{SBAS}{
	short   = SBAS,
	long    = Satellite Based Augmentation Systems,
	}	
	
\DeclareAcronym{IMU}{
	short   = IMU,
	long    = Inertial Measurement Units,
	}

\DeclareAcronym{GNSS}{
	short   = GNSS,
	long    = Global Navigation Satellite System,
	}		

\DeclareAcronym{GPS}{
	short   = GPS,
	long    = Global Positioning System,
	}	
	
\DeclareAcronym{D-GPS}{
	short   = D-GPS,
	long    = Differential \ac{GPS},
	}	
	
\DeclareAcronym{MAP}{
	short   = MAP,
	long    = Maximum A Posteriori,
	}		
	
\DeclareAcronym{EKF}{
	short   = EKF,
	long    = Extended Kalman Filter,
	}	
\DeclareAcronym{UKF}{
	short   = UKF,
	long    = Unscented Kalman Filter,
	}

\DeclareAcronym{BA}{
	short   = BA,
	long    = Bundle-Adjustment,
	}	
		
\DeclareAcronym{DNN}{
	short   = DNN,
	long    = Deep Neural Network,
	}	
	
\DeclareAcronym{GNN}{
	short   = GNN,
	long    = Graph neural networks,
	}	
 
\DeclareAcronym{DL}{
	short   = DL,
	long    = Deep Learning,
	}	
	
\DeclareAcronym{UWB}{
	short   = UWB,
	long    = Ultra Wide Band,
	}	
	
\DeclareAcronym{TLS}{
	short   = TLS,
	long    = Terrestrial Laser Scanner,
	}
	
\DeclareAcronym{MLS}{
	short   = MLS,
	long    = Mobile Laser Scanner,
	}

\DeclareAcronym{SAR}{
	short   = SAR,
	long    = Search and Rescue,
	}	
	
\DeclareAcronym{SfM}{
	short   = SfM,
	long    = Structure from Motion,
	}

\DeclareAcronym{MVS}{
	short   = MVS,
	long    = Multi-View Stereo,
	}

\DeclareAcronym{KPIs}{
	short   = KPIs,
	long    = Key Performance Indicators,
	}

\DeclareAcronym{MVE}{
	short   = MVE,
	long    = Multiview Environment,
	}	
	
\DeclareAcronym{RTPS}{
	short   = RTPS,
	long    = Real Time Positioning System,
	}

\DeclareAcronym{ICP}{
	short   = ICP,
	long    = Iterative Closest Point,
	}

\DeclareAcronym{GICP}{
	short   = GICP,
	long    = Generalized ICP,
	}	
	
\DeclareAcronym{AMCL}{
	short   = AMCL,
	long    = Adaptive Monte Carlo Localization,
	}	
	
\DeclareAcronym{GMCL}{
	short   = GMCL,
	long    = General Monte Carlo Localization,
	}	
	
\DeclareAcronym{SER}{
	short   = SER,
	long    = Similar Energy Region,
	}	
		
\DeclareAcronym{PF}{
	short   = PF,
	long    = Particle Filter,
	}	
	
\DeclareAcronym{RANSAC}{
	short   = RANSAC,
	long    = Random Sample Consensus,
	}	
	
\DeclareAcronym{ROS}{
	short   = ROS,
	long    = Robot Operating System,
	}	

\DeclareAcronym{DoF}{
	short   = DoF,
	long    = Degrees of Freedom,
	}
	
\DeclareAcronym{MAV}{
	short   = MAV,
	long    = Micro Aerial Vehicle,
	}	
	
\DeclareAcronym{VP}{
	short   = VP,
	long    = Vanishing Points,
	}	
	
\DeclareAcronym{VL}{
	short   = VL,
	long    = Vanishing Lines,
	}	
	
\DeclareAcronym{VR}{
	short   = VR,
	long    = Virtual Reality,
	}	
	
\DeclareAcronym{AR}{
	short   = AR,
	long    = Augmented Reality,
	}	
	
\DeclareAcronym{MR}{
	short   = MR,
	long    = Mixed Reality,
	}	
	
\DeclareAcronym{LoD}{
	short   = LoD,
	long    = Level of Detail,
	}	
	
\DeclareAcronym{IFC}{
	short   = IFC,
	long    = Industry Foundation Classes,
	}	
	
\DeclareAcronym{CPS}{
	short   = CPS,
	long    = Cyber-Physical Systems,
	}		
	
\DeclareAcronym{LOAM}{
	short   = LOAM,
	long    = LiDAR Odometry and Mapping,
	}	
	
\DeclareAcronym{A-LOAM}{
  short = A-LOAM,
  long = Advanced implementation of LOAM
  }
  
\DeclareAcronym{F-LOAM}{
  short = F-LOAM,
  long = Fast LiDAR Odometry And Mapping
  }

\DeclareAcronym{IFR}{
  short = IFR,
  long = International Federation of Robotics
  }

\DeclareAcronym{RGB-D}{
  short = RGB-D,
  long = Red-Green-Blue-Depth
  }
   
\DeclareAcronym{OGM}{
  short = OGM,
  long = Occupancy Grid Map
  }
  
\DeclareAcronym{KLD}{
  short = KLD,
  long = Kullback-Leibler distance
  }
    
 \DeclareAcronym{MCL}{
  short = MCL,
  long = Monte Carlo Localization
  }
    
 \DeclareAcronym{GBL}{
  short = GBL,
  long = Graph-based Localization
  }     
  
\DeclareAcronym{RViz}{
  short = RViz,
  long = ROS visualization
  }

\DeclareAcronym{APE}{
  short = APE,
  long = Absolute Pose Error
  }  

\DeclareAcronym{RE}{
  short = RE,
  long = Rotational Error 
  } 
  
  \DeclareAcronym{RMSE}{
  short = RMSE,
  long =  Root Mean Square Error 
  } 
  
  \DeclareAcronym{PGBM}{
  short = PGBM,
  long =  Pose Graph-based Maps 
  } 

 \DeclareAcronym{CAD}{
  short = CAD,
  long =  Computer-aided Design
  } 

 \DeclareAcronym{Ogm2Pgbm}{
  short = Ogm2Pgbm,
  long =  \ac{OGM} to Pose Graph-based map
  } 

 \DeclareAcronym{Scan-BIM deviations}{
  short = Scan-BIM deviations,
  long =  discrepancies between the reference \ac{BIM model} and the real world
  } 
  
 \DeclareAcronym{SC}{
  short = SC,
  long =  Scan Context
  } 
  
\DeclareAcronym{PD}{
  short = PD,
  long =  Positive Differences
  }

\DeclareAcronym{ND}{
  short = ND,
  long =   Negative Differences
  } 

\DeclareAcronym{VG}{
  short = VG,
  long =  Voxel Grid
  } 
  
\DeclareAcronym{SD}{
  short = SD,
  long =  Session Data
  } 
  
\DeclareAcronym{MSS}{
  short = MSS,
  long =  multi-session SLAM
  } 
  
\DeclareAcronym{PP}{
  short = PP,
  long =  Path Planner
  } 
  
\DeclareAcronym{AEC}{
  short = AEC,
  long =  {Architecture, Engineering and Construction} 
  }

\begin{document}

\linespread{0.5}

\title{BIM-SLAM: Integrating BIM Models in Multi-session SLAM for Lifelong Mapping using 3D LiDAR}


\author{M. A. Vega Torres$^{1}$ 
\& A. Braun$^{1}$ 
\& A. Borrmann$^{1}$}

\affiliation{
$^1$Chair of Computational Modeling and Simulation, Technical University of Munich, Munich, Germany
}

\email{
\href{mailto:e.miguel.vega@tum.de}{miguel.vega@tum.de},
\href{mailto:alex.braun@tum.de}{alex.braun@tum.de},
\href{mailto:andre.borrmann@tum.de}{andre.borrmann@tum.de}
}

\maketitle 
\thispagestyle{fancy} 
\pagestyle{fancy}

\begin{abstract}
While 3D \ac{LiDAR} sensor technology is becoming more advanced and cheaper every day, the growth of digitalization in the \ac{AEC} industry contributes to the fact that 3D building information models (\acs{BIM model}s) are now available for a large part of the built environment. 
These two facts open the question of how 3D models can support 3D \ac{LiDAR} long-term \ac{SLAM} in indoor, \ac{GPS}-denied environments.
This paper proposes a methodology that leverages \ac{BIM model}s to create an updated map of indoor environments with sequential \ac{LiDAR} measurements. 
Session data (pose graph-based map and descriptors) are initially generated from \ac{BIM model}s.
Then, real-world data is aligned with the session data from the model using multi-session anchoring while minimizing the drift on the real-world data.
Finally, the new elements not present in the \ac{BIM model} are identified, grouped, and reconstructed in a surface representation, allowing a better visualization next to the \ac{BIM model}.
The framework enables the creation of a coherent map aligned with the \ac{BIM model} that does not require prior knowledge of the initial pose of the robot, and it does not need to be inside the map.
\end{abstract}


\section{Introduction}
\label{sec:Introduction}

The ability to align, compare, and manage data acquired at different times that could be spaced apart by long periods, also known as long-term map management, is crucial in real-world robotic applications.

Since the real world is permanently evolving and changing, long-term map management is essential for autonomous robot navigation and users who want to use the map to understand the current situation and its evolution.
In particular, in an emergency, an up-to-date map can serve first responders to increase situational awareness and support decision-making to save lives efficiently and safely \cite{alliez2020real}. 



Maps are usually created with mobile robots equipped with sensors and leveraging \ac{SLAM} algorithms to enable fast and automated workflows.
However, these maps are commonly disconnected from any preliminary information, creating a map that may suffer from significant drift and does not allow change detection or comparison with a prior map.
However, these maps are commonly disconnected from any preliminary information, creating a map that may suffer from significant drift and does not allow change detection or comparison with a prior map.

Nonetheless, a georeferenced \ac{BIM model} is available for most contemporary buildings and can be used as a reference map to enable accurate \ac{LiDAR} localization and mapping.

In addition, the robot's pose in the coordinate system of the \ac{BIM model} could be retrieved with a localization algorithm \cite{vega:2022:2DLidarLocalization}. 
Given this pose, the \ac{BIM model} could support autonomous robotic tasks. 
For example, path planning, object inspection \cite{Kim.2022} or maintenance and repair \cite{Kim.2021}. 

As will be discussed in Section \ref{Chap:related_work}, several researchers have investigated the use of \ac{BIM} for robot localization \cite{vega:2022:2DLidarLocalization}. 
However, only a few aim to create an accurate, updated map aligned with the information from the \ac{BIM model}. 

Furthermore, most of them also require a perfect estimation of the robot's initial position, which must be inside the prior map.
On top of that, almost no method considers \ac{Scan-BIM deviations}. 
While we allow \ac{Scan-BIM deviations}, we also assume that the model still represents a reliable map suitable for localization, i.e., the BIM contains enough features that coincide geometrically with the real world.

To address this challenge, we propose a novel framework that allows the improvement of a map created with mobile 3D LiDAR data. The map is corrected as it is aligned with a BIM model, allowing a better understanding of the scene and change detection for long-term map management.

First, we create session data from \ac{BIM model}s. 
Session data (SD) represent data collected from the exact location at various periods. 
These data are very convenient for performing offline operations between sessions, e.g., inter-session alignment \cite{kim2022ltMapper} or place recognition \cite{kim2022ScLidar}. 

A proposed ground-truth multi-session anchoring uses these \ac{SD} to align and rectify another query \ac{SD} with a reference \ac{BIM model}.

Finally, the aligned data is compared against the \ac{BIM model}, and positive differences are represented as reconstructed surfaces, enabling a better understanding of the current environment.

Overall, the proposed technique aims to contribute to accurately mapping indoor \ac{GPS}-denied environments and posterior long-term map management.

The remainder of this paper is organized as follows.
Section \ref{Chap:related_work} describes related work on BIM-based LiDAR localization and mapping. 
Section \ref{chap:methodology} introduces our modular \textbf{BIM-SLAM} framework, which is divided into three main steps: \textbf{1.} Generation of \ac{SD} from \ac{BIM model}s, \textbf{2.} Alignment of new session data with the reference \ac{BIM model} and \textbf{3.} Positive change detection and segmentation of new elements.
Section \ref{chap:experiments} presents the experimental settings and implementation details. 
Finally, section \ref{chap:conclusions} summarizes our contributions and concludes our work.

\section{Related research}
\label{Chap:related_work}

Besides having the potential to support robot localization, a 3D \ac{BIM model} can be used as a prior map to create accurate, updated 3D maps and allow fast autonomous navigation.

This section will overview state-of-the-art methods, which used prior building information, i.e., \ac{BIM model}s or floor plans, to support robot localization and also methods that support mapping.

\subsection{BIM-based 2D \ac{LiDAR} localization and mapping}

\citeauthor{Follini.2020}\cite{Follini.2020} demonstrate how the standard \ac{AMCL} algorithm may be used to obtain the transformation matrix between the robot's reference system and a map that was extracted from the \ac{BIM model}. 

The same algorithm was used by \cite{Prieto.2020}, \cite{Kim.2021}, and \cite{Kim.2022} to localize a wheeled robot in an \ac{OGM} generated from the \ac{BIM model}. The main difference between these approaches relies on how they create the \ac{OGM} from the \ac{BIM model}. 

\citet{Hendrikx.2021} proposed an approach that, instead of using an \ac{OGM}, applies a robot-specific world model representation extracted from an \ac{IFC} file for 2D-\ac{LiDAR} localization. 
In their factor graph-based localization approach, the system queries semantic objects in its surroundings (lines, corners, and circles) and creates data associations between them and the laser measurements. 
More recently, in \cite{hendrikx2022local}, they improved and evaluated their method for global localization, achieving better results against \ac{AMCL}.

Instead of using a \ac{BIM model}, \cite{Boniardi.2017} use a CAD-based architectural floor plan for 2D \ac{LiDAR}-based localization. In their localization system, they implement \ac{GICP} for scan matching with a pose graph \ac{SLAM} system. 
Later, they proposed an improved pipeline for long-term localization and mapping in dynamic environments performing better than \ac{MCL} in the pose tracking problem \cite{Boniardi.2019}.

In our previous work \cite{vega:2022:2DLidarLocalization}, besides proposing a method to create an \ac{OGM} from a multi-story \ac{IFC} Model, we demonstrated that the widely used \ac{AMCL} is not that robust to deal with changing and dynamic environments, as compared to \ac{GBL} methods, like Cartographer and SLAM Toolbox.
Based on these findings, and in an effort to facilitate the transition from \ac{PF} to \ac{GBL} methods, we also contributed with an open source method that converts \ac{OGM} to \ac{PGBM} for robust robot pose tracking. \cite{vega:2022:2DLidarLocalization}.

\subsection{BIM-based 3D \ac{LiDAR} localization and mapping}

Other methods have explored 3D \ac{LiDAR} localization with \ac{BIM model}s.

A high-accuracy robotic building construction system was proposed by \cite{gawel2019fully}. Besides a robust state estimator that fuses \ac{IMU}, 3D \ac{LiDAR}, and Wheel encoders, they use ray-tracing with three \textit{laser-distance sensors} and a 3D \ac{CAD} model to localize the end-effector with sub-cm accuracy. To accomplish this, they took several orthogonal range measurements while the robot was stationary.

In \cite{ErcanJenny.2020} and \cite{Blum.2020}, the 3D \ac{LiDAR} scan is aligned with the \ac{BIM model} using the \ac{ICP} algorithm. 
Whereas in \cite{ErcanJenny.2020}, the alignment is constrained to a few selected reference-mesh faces to overcome ambiguities, \cite{Blum.2020} use image information to filter foreground and background in the point cloud and use only the background for registration. 
Later the pipeline was reconfigured to create a self-improving semantic perception method that can better handle clutter in the environment \cite{Blum.2021}.

\begin{figure*}[!ht]
\centering{
\def\svgwidth{\textwidth}
{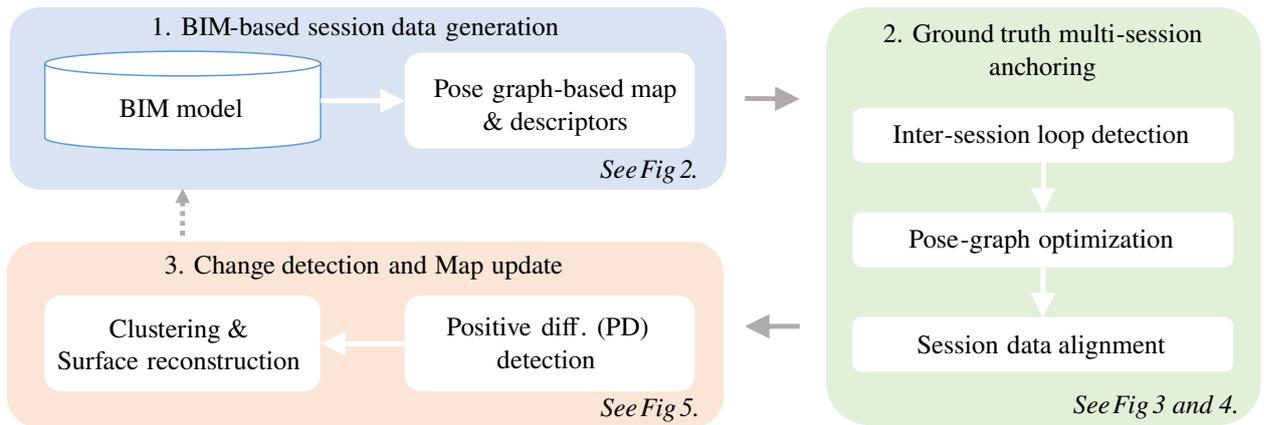}
\caption{Overview of \textbf{BIM-SLAM}. The pipeline consists of three steps: BIM-based session data generation, Multi-session anchoring, and Map update.}
\label{fig:method}
}
\end{figure*}

\cite{Moura2021} propose a technique to create \textit{.pbstream} maps from \ac{BIM model}s and achieve localization using Cartographer. 
This method is very convenient; however, since they use Cartographer in localization mode, the robot must have its initial pose inside the boundaries of the prior map to be localized and create an aligned map.

\cite{Oelsch2021} propose Reference-LOAM (R-LOAM), a method that leverages a joint optimization incorporating point and mesh features for 6 degrees of freedom (DoF) \ac{UAV} localization. Subsequently, in \cite{Oelsch.2022}, they improved their method with pose-graph optimization to reduce drift even when the reference object is not visible. 

Recently, \cite{YIN_2023_104641} introduced a semantic \ac{ICP} method that can leverage the 3D geometry and the semantic information of a \ac{BIM model} for data associations achieving a robust 3D \ac{LiDAR} localization method. Their framework proposes a BIM-to-Map conversion, converting the 3D model into a semantically enriched point cloud. 
Their experiments show that they can achieve effective 3D \ac{LiDAR}-only localization with a \ac{BIM model}.

 \cite{shaheer2022robot} proposed another relevant approach; here, instead of using object semantics for localization, they use geometric and topological information in the form of walls and rooms. With this information, they create Situational Graphs (S-Graphs), which are then used for accurate pose tracking.
Later, they enhanced their method allowing the creation of a map prior to localization and posterior matching and merging with an A-graph (extracted from \ac{BIM model}s). The final merged map was denoted as an informed Situational Graph (iS-Graph) \cite{shaheer2023graphbased}.

In \cite{caballero2021dll}, the authors introduced direct \ac{LiDAR} localization (DLL), a fast direct 3D point cloud-based localization technique using 3D \ac{LiDAR}. 
They employ a registration technique that does not require features or point correspondences and is based on non-linear optimization of the distance between the points and the map.
The approach can track the robot's pose with sub-decimeter accuracy by rectifying the expected pose from odometry. The method demonstrated better performance than AMCL 3D. 


While numerous strategies aiming to leverage \ac{BIM model}s for  \ac{LiDAR} localization and mapping have emerged. 
Most of them have focused on real-time localization without allowing a better estimation of previous poses with pose-graph-based optimization techniques.

Moreover, almost all must have a good initial pose inside the given map.
Without this initial pose and if the robot starts from a point where the \ac{BIM model} is not visible, there is no possibility for localization or the creation of an aligned map.
Furthermore, most focused on something different than automatically identifying the environmental discrepancies.

In this paper, we propose a method that handles these issues, demonstrating that it is possible to retrieve an accurate, aligned, updated, optimized map close to the ground truth and identify positive differences. 
The method also works if the robot's starting position is not inside the map.

\section{Methodology}
\label{chap:methodology}

\subsection{Overview}

Our method can be divided into three main steps, as illustrated in Figure \ref{fig:method}:
\textbf{Step 1:} Automatic generation of ground truth \ac{SD} from an \ac{IFC} model employing \ac{Ogm2Pgbm} \cite{vega:2022:2DLidarLocalization} and Gazebo simulator. 
\textbf{Step 2:} Multi-session anchoring with a \ac{BIM model} as ground truth. 

\textbf{Step 3:} 3D aligned map construction and change detection.

\begin{figure}[!htb]
    \centering
    \includegraphics[width=0.48\textwidth]{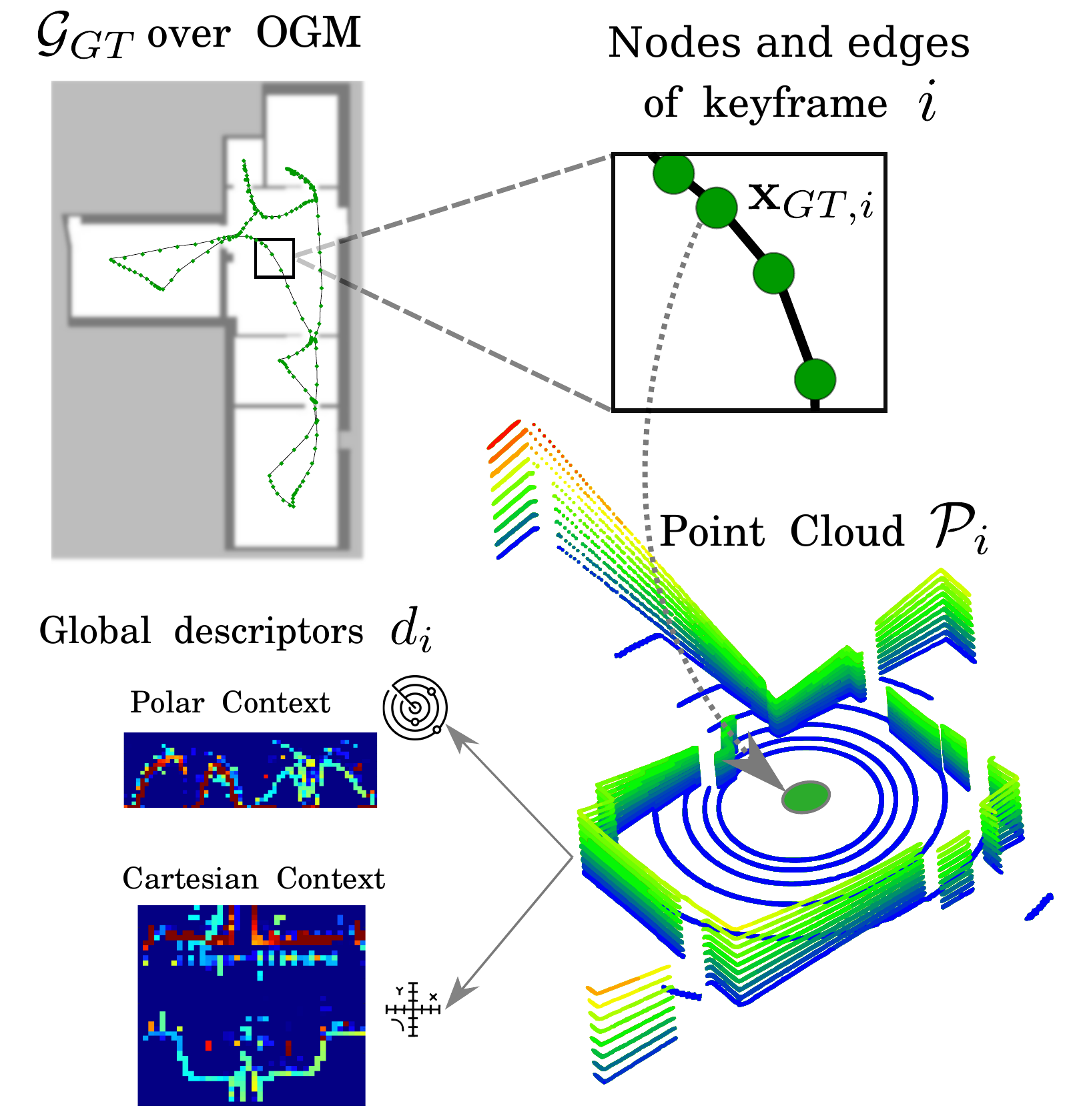}
    \caption{Generated \ac{SD} from the \ac{BIM model}. In the top left, the ground truth pose graph-based map $\mathcal{G}_{GT}$, with its nodes and edges (the result of odometry constraints). Besides having position and orientation, each node $\mathbf{x}_{GT, i}$ has its corresponding laser scan $\mathcal{P}_i$ and global descriptors  $d_i$. In this research, we use only polar context descriptors.}
    \label{fig:sessionData}
\end{figure}

\subsection{BIM-based session data generation}
Using parts of our previous contribution \cite{vega:2022:2DLidarLocalization}, 3D \ac{LiDAR} \ac{SD} with ground truth poses can be created from the \ac{BIM model}.
As a first step, an \ac{OGM} is created from the model, which is then used to determine the path where a robot will be simulated.
The generation of \ac{SD} from an \ac{OGM} is performed with the open source \ac{Ogm2Pgbm} package \cite{vega:2022:2DLidarLocalization}, with a slight modification. 
Instead of simulating 2D laser scan \ac{ROS} messages with ray casting, we leverage the Gazebo physics simulation engine \cite{Gazebo_koenig2004design} to simulate the complete 3D \ac{LiDAR} scans.
This simulation is done by sending navigational goals to a simulated robot and leaving it to navigate autonomously with the \ac{ROS} Navigation Stack through the estimated path provided by the Wavefront Coverage \ac{PP} of \ac{Ogm2Pgbm}. 

The navigational goals are found by retrieving only one waypoint out of 20 in the list of generated waypoints from the \ac{PP}. 
In this way, the waypoints are separated by a considerable distance but still allow the robot to navigate the complete path without skipping any region.
Then the selected waypoints are passed as navigational goals (with pose and orientation) to Gazebo only when the simulated robot is at a certain distance to the next selected waypoint.
These navigational goals are necessary to allow the robot to execute the whole exploration smoothly without unnecessary stops or skipping previous navigational goals.

It is necessary to recall that in this step, the semantics of the \ac{BIM model} are leveraged, extracting only permanent structures (for example, walls, ceilings, and floors) and excluding spaces, non-permanent or \ac{LiDAR}-invisible objects (like doors, windows, or curtain walls).
Please, refer to \cite{vega:2022:2DLidarLocalization} for more details about this step.  

Once the 3D \ac{LiDAR} data is simulated with ground truth odometry, we implement a \textit{keyframe information saver} \cite{kim2022ltMapper} to generate \textit{session data} from it. A session $\mathcal{S}$ is defined as

\begin{equation}\label{eq:1}
\mathcal{S}:=\left(\mathcal{G},\left\{\left(\mathcal{P}_i, d_i\right)\right\}_{i=1, \ldots, n}\right) ,
\end{equation}

where $\mathcal{G}$ is a pose-graph text file with pose nodes, odometry edges, and optionally recognized intra-session loop edges. 

The $\left(\mathcal{P}_i, d_i\right)$ are the 3D point cloud and the global descriptor of the $i^{th}$ keyframe, and $n$ is the total number of equidistantly sampled keyframes. Figure \ref{fig:sessionData} illustrates the elements of the \ac{SD}.


\subsection{Ground-truth multi-session anchoring}


Given the \ac{SD} from the real-world data $\mathcal{S_Q}$, denoted as \textit{query}, and the \ac{SD} from the \ac{BIM model} $\mathcal{S_{GT}}$, denoted as \textit{ground truth session}, our goal is then to align $\mathcal{S_Q}$ with $\mathcal{S_{GT}}$ and retrieve a globally consistent map.

To align these maps, we use anchor node-based inter-session loop factors, as described in \cite{kim2022ltMapper}, \cite{kim2010multiple}, \cite{McDonald2013_1144}, \cite{ozog2016long}.
Instead of optimizing the poses of the two sessions, we do not allow modifications of the ground truth session created from the model. 
This step is based on the assumption that, while there can be \ac{Scan-BIM deviations}, the model still represents a reliable map suitable for localization.

In order to avoid alterations to the current session in $\mathcal{S_{GT}}$, we add its poses as \textit{prior factors} with a very low variance (i.e., \SI{1e-100}{\nothing}) in its noise model, to the pose-graph optimization problem.

\textit{Anchor nodes} represent the transformation from the local frame of each session graph to a standard global reference frame. 
Once the poses of each session are transformed into the global frame (via the anchor nodes), a comparison of the measurements between sessions is possible.

As outlined in \citet{kim2010multiple}, the anchor node can successfully approximate the offset between sessions. Besides allowing faster convergence to the least-squares solvers, anchor nodes allow each session to optimize their poses before global constraints are observed \cite{ozog2016long}. 

This characteristic is beneficial for long-term mapping since it allows the creation of a first consistent map in the state of the environment at the time the data was collected. Once a map is created with a new session, the anchor nodes allow finding the transformation that aligns that session with another reference one. 
In our case, the reference session is extracted from the \ac{BIM model}.

To add the anchor node to the pose graph optimization problem, we only need to add the following factor:

\begin{equation}\label{eq:2}
\begin{aligned}
& \phi\left(\mathbf{x}_{Q, j}, \Delta_Q\right) \\
& \propto \exp \left(-\frac{1}{2}\left\|\left(\left(\Delta_{GT} \oplus \mathbf{x}_{GT, i}\right) \ominus\left(\Delta_Q \oplus \mathbf{x}_{Q, j}\right)\right)-c\right\|_{\Sigma_c}^2\right),
\end{aligned}
\end{equation}

Here $\mathbf{x}$ is a SE(3) pose; $c$ is an encounter \cite{kim2010multiple}, $i$ and $j$ are pose indexes; $\oplus$ and $\ominus$ are the SE(3) pose composition operators.
$\Delta$ indicates an anchor node, which is also a SE(3) pose variable. 
Whereas the query session's anchor node $\Delta^Q$ has a relatively high covariance, the ground truth's  $\Delta^{GT}$ has an insignificant value (very close to zero). 


\begin{figure}[!htb]
    \centering
    \includegraphics[width=0.48\textwidth]{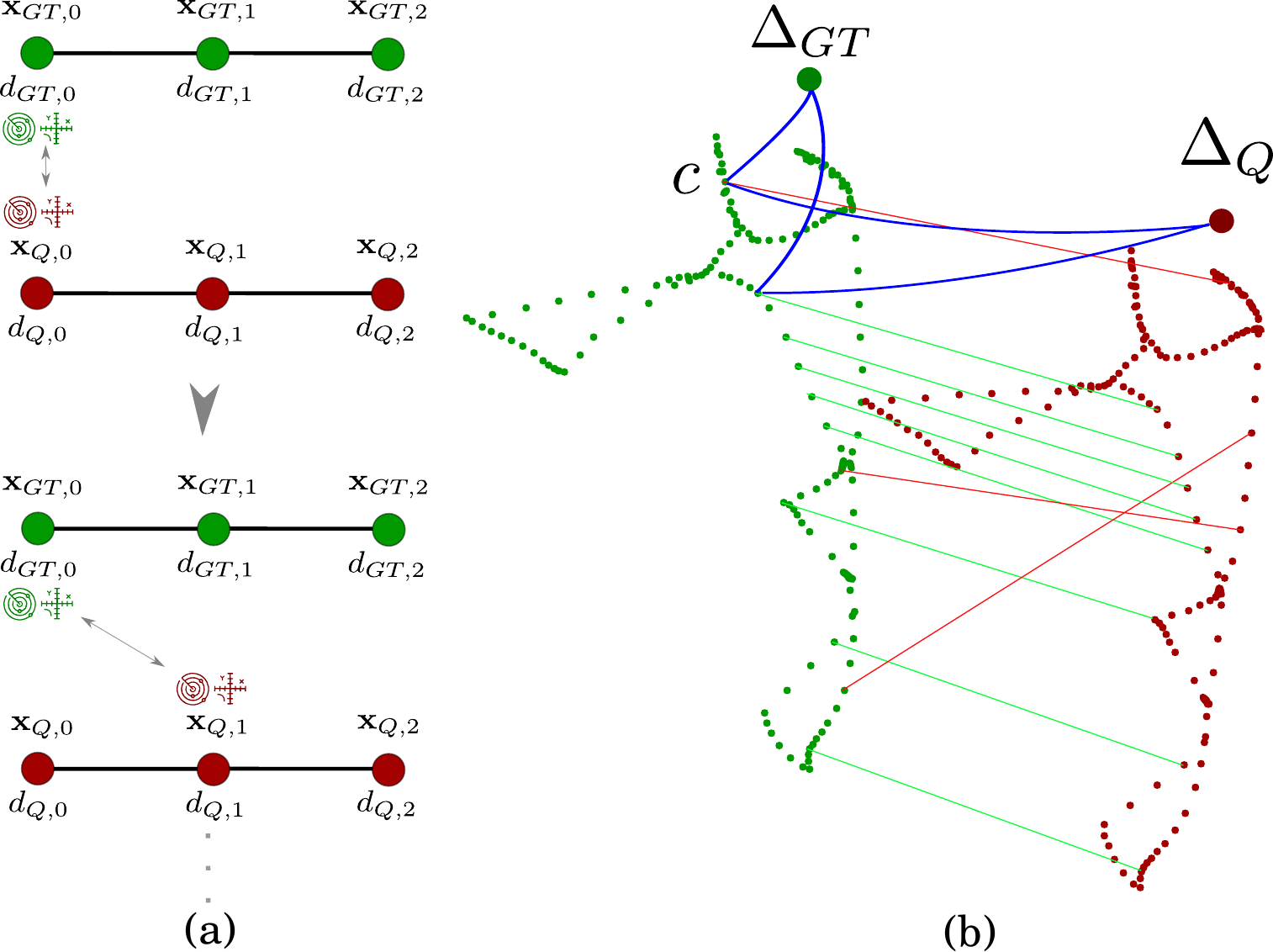}
    \caption{
    Loop closure detection between sessions. (a) the descriptors of each session data are compared against each other to find correspondences. (b) Some correctly and wrongly detected loop closures are shown in green and red, respectively. In blue, two encounters $c$ linked to the trajectories' respective anchors. 
    The trajectory's offset concerning a common global frame is specified by the anchors $\Delta$. 
    }
    \label{fig:LC}
\end{figure}
\nointerlineskip

To identify loop-closure candidates between sessions, we use \ac{SC} \cite{kim2018scan} and the pose proximity-based radius search loop detection. Figure \ref{fig:LC} exemplifies the inter-session loop closure detection process. 
Once loops are detected, a 6D relative constraint between two keyframes is determined by registering their respective laser scan point clouds  $\mathcal{P}_{GT, i}$ and $\mathcal{P}_{Q, j}$ with \ac{ICP}.
As in \cite{kim2022ltMapper}, only loops with adequate low \ac{ICP} fitness ratings are allowed, and the score is used to calculate an adaptive covariance $\Sigma_c$ in (\ref{eq:2}).

\begin{figure}[!htb]
    \centering
    \includegraphics[width=0.48\textwidth]{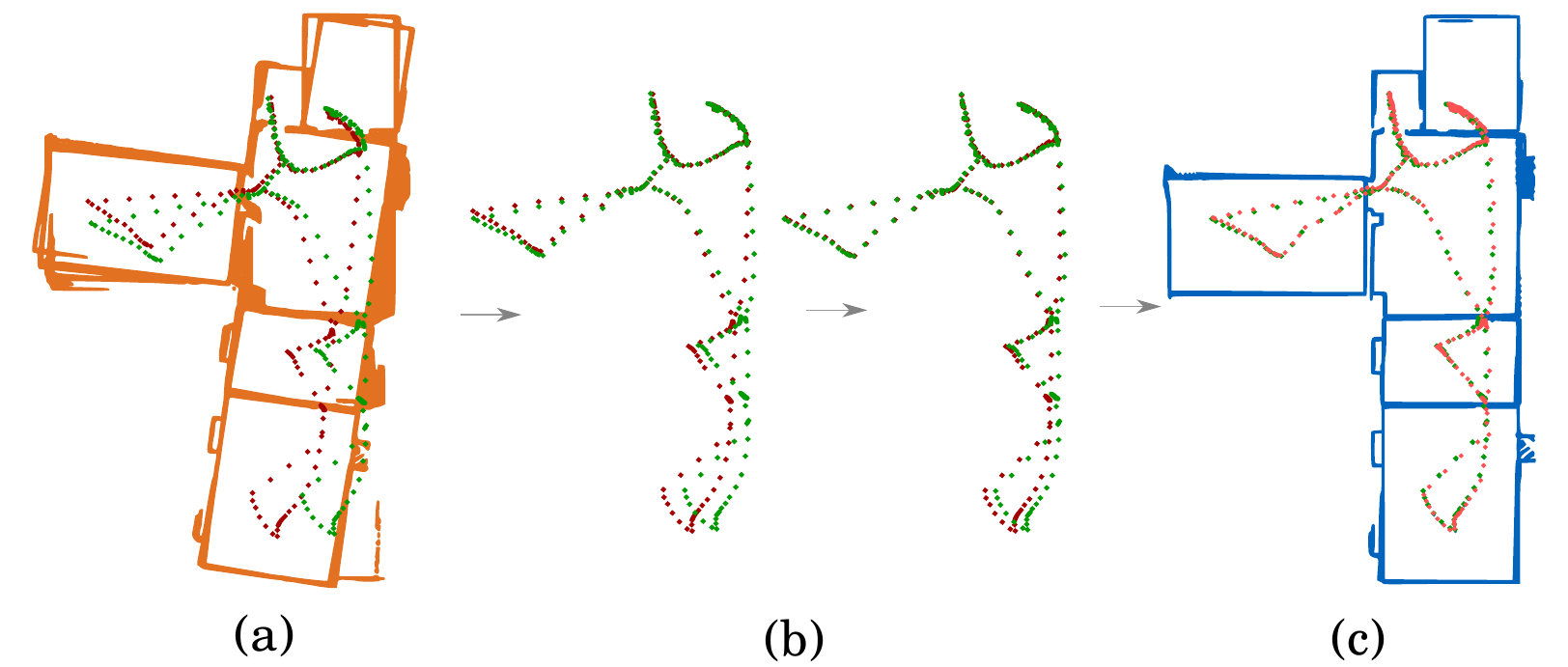}
    \caption{Pose graph optimization with multiple sessions. (a) In orange is the drifted map created by a SLAM system (exaggerated for illustrative purposes); in red is the respective trajectory; in green is the ground truth trajectory (b) Each pose graph optimization iteration tries to create a consistent global map placing the scans closer to the ground truth. (c) The final map is correctly aligned with the \ac{BIM model}. 
    }
    \label{fig:PGO}
\end{figure}

Once the values of the anchor nodes and the poses on the local coordinate system of $\mathcal{S_{Q}}$ are optimized, as shown in Figure \ref{fig:PGO}, all the poses in the pose-graph can be converted from the local (denoted as ${ }^Q \mathcal{G}_Q^*$ ) to the global coordinate system ${ }^W \mathcal{G}_Q^*$ by applying the following transformation to each pose $\mathbf{x}$ in a graph:
$$
{ }^W \mathbf{x}_Q^*=\Delta_Q^* \oplus^Q \mathbf{x}_Q^* ,
$$
where $W$ is the global coordinate system.

\setlength{\belowcaptionskip}{-20pt}
\nointerlineskip

\subsection{Aligned map construction and Change detection}


With the calculated poses on the same coordinate system, a map can be created by placing the respective laser scans $\mathcal{P}_{Q, i}$ in the estimated poses ${ }^W \mathbf{x}_{Q, i}^*$, which are now in the BIM coordinate system.

Once the 3D map of the current state of the environment is aligned with the \ac{BIM model}, a comparison of the two maps is possible.
A distance threshold is set to differentiate between objects present in the model and the new objects in the updated environment, also known as \textit{\ac{PD}}.

A signed distance computation lets us determine which points are close to and far from the mesh. 
Close points allow confirming \ac{BIM model} elements, and far points are considered new elements, i.e., elements not present in the model.

In the next step, a  density-based clustering algorithm (DBSCAN) is used to split the point cloud of detected \ac{PD}s into segments of points that are close to each other and might represent single objects. 
This results in better visualization of the new objects in combination with the model, allowing a better scene understanding.


Finally, each cluster of the \ac{PD}s is converted to a mesh representation created using cubes from a \ac{VG} of the point cloud. 

\begin{figure}[!htb]
    \centering
    \includegraphics[width=0.48\textwidth]{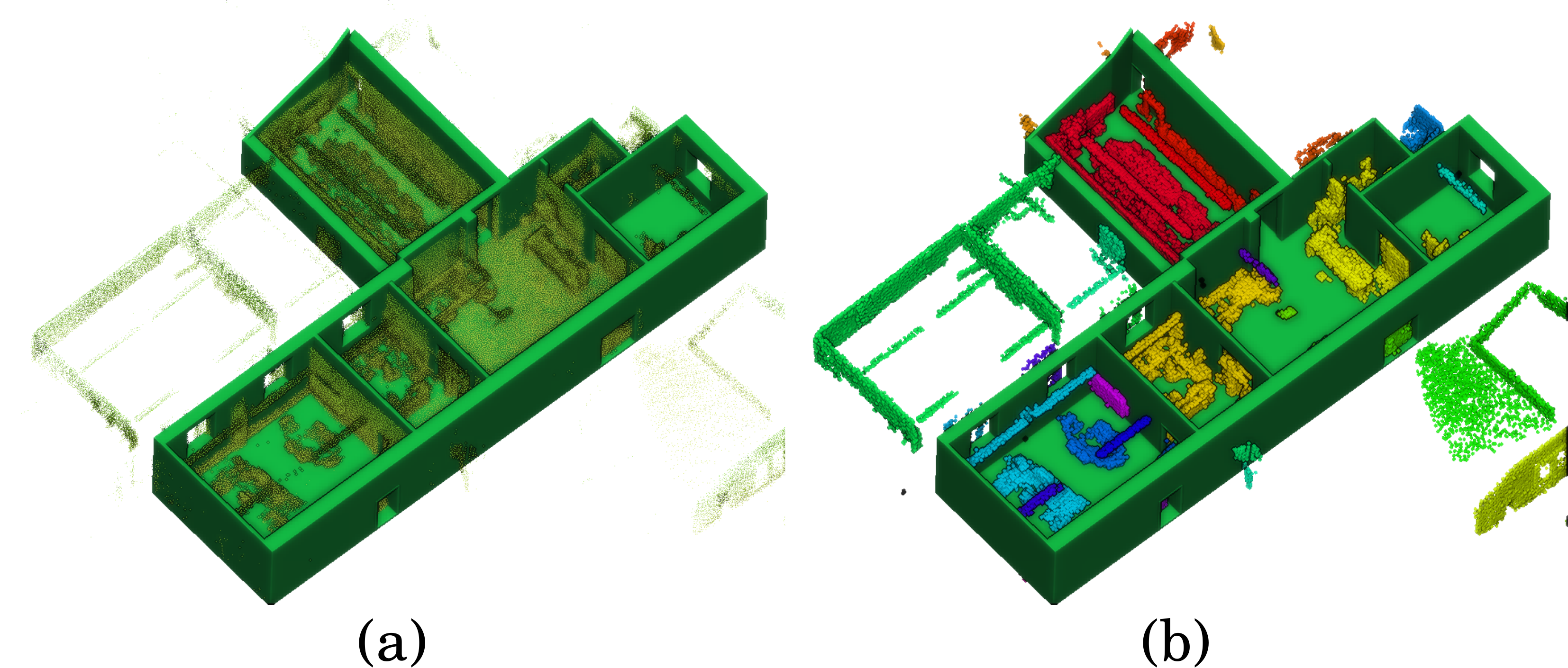}
    \caption{Detected positive differences in the point cloud and the \ac{BIM model}. (a) The original aligned corrected point cloud (b) Only voxelized clustered new objects in the scene. The ceiling was removed for better visualization.}
    \label{fig:final_result}
\end{figure}

Compared with other surface reconstruction methods, voxels represent the actual geometry of the objects visible in the scene.
A final result is visible on \ref{fig:final_result}.

\nointerlineskip

\section{Experiments and results}
\label{chap:experiments}

The data used for evaluating the proposed strategies are presented in this section, together with implementation and evaluation details. Both simulated and real environments were used for the experiments.

\subsection{Simulated experiments}
We used Gazebo \cite{Gazebo_koenig2004design} to simulate the experimental data.

The robot used for the simulated experiments was the Robotnik SUMMIT XL, equipped with a Velodyne VLP-16 3D LiDAR. 
Table \ref{tab:results} presents the results on the simulated sequence.

\vspace{-0.5cm}
\textcolor{blue}{
\begin{table}[htbp]
  \centering
  \caption{Quantitative comparative results.}
    \begin{tabular}{c|c c|c c}
    \toprule
   \multirow{2}{*}{Method}  & \multicolumn{2}{c|}{Trans. Error (cm)} & \multicolumn{2}{c}{Rot. Error (deg)} \\
   ~ & RMSE & Max & RMSE & Max \\ \hline
     SC-A-LOAM & 12,439 & 24,015 & 2,316 & 5,276 \\ 
    \textit{BIM-SLAM} & 10,594 & 20,175 & 2,074 & 4,841 \\ 
    \bottomrule
    \end{tabular}%
  \label{tab:results}%
\end{table}%
}

\vspace{-0.5cm}

\subsection{Real-World experiments}
\begin{figure}[!htb]
    \centering
    \includegraphics[width=0.4\textwidth]{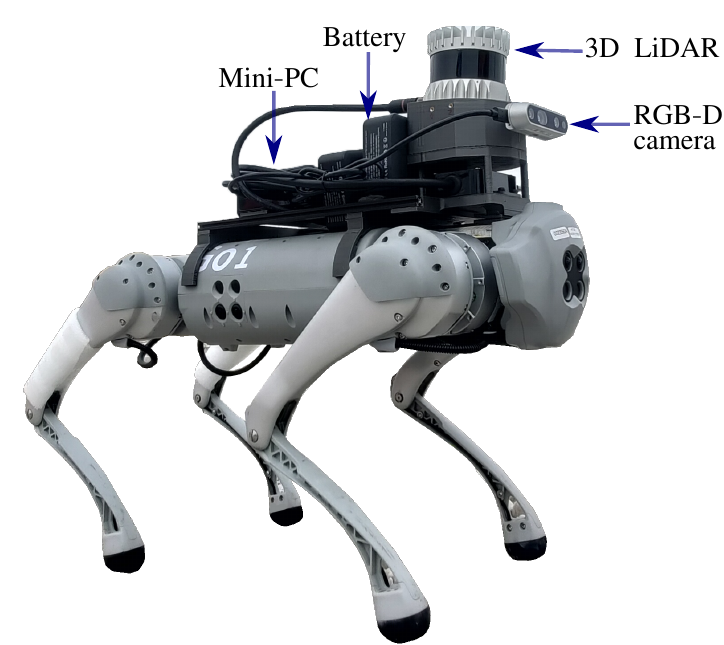}
    \caption{Portable mapping system developed to acquire real-world data. Here it is placed over the Go1 quadruped robot.}
    \label{fig:robotAndMappingSys}
\end{figure}

The real-world data was collected in the same environment as the simulated one with an Ouster OS1-32 LiDAR.
The sensor was mounted with a mini-PC and batteries in a mapping system.
This system can be used as a handheld or above a robotic platform as depicted in \ref{fig:robotAndMappingSys}.
The legged robot Go1 was the robot platform we utilized for these tests. 
This data was used to generate the results shown in Figure \ref{fig:final_result}.

\nointerlineskip

\subsection{Implementation details}
While the \textit{Step 1} and \textit{3} were implemented in Python, \textit{Step 2} was written in C++ using ROS noetic.

Ogm2Pgbm \cite{vega:2022:2DLidarLocalization}, Scan Context \cite{kim2018scan}, and LT-SLAM \cite{kim2022ltMapper} were sources that we used that are freely available online. The parameters of \ac{SC} were adjusted for indoor environments; for example, we use a maximum radius of \SI{10}{\meter} and a cosine similarity threshold between descriptors of \num{0.6}.

Whereas in \textit{Step 2}, the pose-graph optimization is done with GTSAM using  iSAM2, in \textit{Step 3}, the process is done with Trimesh and Open3D. 

The generation of the \ac{SD} was done with SC-A-LOAM \cite{kim2022ScLidar} an enhanced version of A-LOAM with loop closure and key-frame information saving capabilities (i.e., for \ac{SD} generation). 

\section{Discussion}

While in terms of drift reduction, our method performs only slightly better than state-of-the-art \ac{SLAM} algorithms, as shown in Table \ref{tab:results}.
The main advantage of the proposed technique relies on the alignment of the acquired sensor data with the \ac{BIM model}. 
The method also allows the creation of a map without the sensor being inside the prior map (as localization algorithms would require) or knowing its initial position.

We also show how the aligned sensor data can improve situational awareness (see Figure \ref{fig:final_result}). Moreover, once the data is aligned, the model can be leveraged to identify new and missing components and clean the point cloud, removing unnecessary noisy points from the surroundings. These features would enable robust long-term map data management, using less memory than needed and saving all the information in one reference coordinate system.

However, our pipeline still presents some limitations. For example, in the case of significant levels of \ac{Scan-BIM deviations}, the minimal overlap between the scan and the BIM, or symmetric environments, the correct alignment may not be possible with our method. 
Adjustments in creating the point cloud descriptors or the corresponding matching process might be necessary to address this issue.

Nonetheless, the method is not restricted to Manhattan-world environments with enclosed rooms, like in \cite{shaheer2023graphbased}, nor does it require manual input of the robot's initial position in the map like the one proposed in \citep{Oelsch.2022}.

Avenues for future research include enhancements in the place recognition algorithm (here, we use \ac{SC}); leveraging topological and semantic information from the \ac{BIM model} to make most robust the alignment and optimization process; and improvements on the extraction of \ac{SD} from \ac{BIM model}s, in terms of speed and feature extraction, perhaps with a faster rendering method.

\section{Conclusions}
\label{chap:conclusions}

This paper presents a modular pipeline to allow 3D \ac{LiDAR} data alignment and change detection with a \ac{BIM model} as a reference map. Contrary to several other approaches, we aim to create an accurate, consistent map of the current state rather than focusing on real-time performance.
The method does not need to know the robot's initial position, nor does the robot need to start inside the given map.
In this way, our framework allows map alignment and extension even if the reference \ac{BIM model} is no longer visible or if only a part of the model is scanned.

In the future, we would like to test the method on \ac{SD} with more environmental changes and to enhance the change detection module to handle negative differences, i.e., when parts of the original \ac{BIM model} are no longer present in the environment.

\section{Acknowledgement}
The presented research was conducted in the frame of the project ``Intelligent Toolkit for Reconnaissance and assessmEnt in Perilous Incidents'' (INTREPID) funded by the EU's research and innovation funding programme Horizon 2020 under Grant agreement ID: 883345.

\bibliography{03_References/references}

\end{document}